\documentclass{article}

\usepackage[utf8]{inputenc}

\usepackage{amsfonts} 
\usepackage{graphicx}
\usepackage{pgfplots} % LaTeX
\usepackage{tikz}
\usepackage{mathrsfs}
\usepackage{amssymb}
\usepackage{graphicx}
\usepackage{amsmath}
\usepackage{latexsym}
\usepackage{amsthm}
\usepackage{subfig}
\usepackage{arxiv}
\usepackage{url}
\title{Data Exploration and Validation on dense knowledge graphs for biomedical research}
\date{April 2019}
\author{Jens Dörpinghaus\footnote{Fraunhofer Institute for Algorithms and Scientific Computing,
Schloss Birlinghoven, Sankt Augustin, Germany  \url{jens.doerpinghaus@scai.fraunhofer.de}} \and
Alexander Apke\footnote{Department for Mathematics and Computer Science,
University of Cologne, Germany \url{apke@zpr.uni-koeln.de}} \and
Vanessa Lage-Rupprecht\footnote{Fraunhofer Institute for Algorithms and Scientific Computing,
Schloss Birlinghoven, Sankt Augustin, Germany  \url{vanessa.lage-rupprecht.fraunhofer.de}} \and
Andreas Stefan\footnote{Fraunhofer Institute for Algorithms and Scientific Computing,
Schloss Birlinghoven, Sankt Augustin, Germany  \url{andreas.stefan@scai.fraunhofer.de}} }
\begin{document}
\maketitle

\begin{abstract}
Here we present a holistic approach for data exploration on dense knowledge graphs as a novel approach with a proof-of-concept in biomedical research.
Knowledge graphs are increasingly becoming a vital factor in knowledge mining and discovery as they connect data using technologies from the semantic web.
In this paper we extend a basic knowledge graph extracted from biomedical literature by context data like named entities and relations obtained by text mining and other linked data sources like ontologies and databases. We will present an overview about this novel network. The aim of this work was to extend this current knowledge with approaches from graph theory.
This method will build the foundation for quality control, validation of hypothesis, detection of missing data and time series analysis of biomedical knowledge in general. In this context we tried to apply multiple-valued decision diagrams to these questions. In addition this knowledge representation of linked data can be used as FAIR approach to answer semantic questions.
%Beside of applications of this approach for life sciences and personalized medicine we will also describe problems for data mining (missing and uncertain data)
This paper seeks to address technological problem and especially the storage and analysis of this data. %we will show that data validation and exploration can be done using multi-valued decision diagrams.
This paper sheds new lights on dense and very large knowledge graphs and the importance of a graph-theoretic understanding of these networks.
\end{abstract}
%\begin{keywords}
%Text mining  \and Natural language processing \and Network science \and Linked data \and Knowledge representation \and Semantic query languages %Keyword1 \and Keyword2
%\end{keywords}
%%% Beginn des Artikeltexts
\section{Introduction}

Here we present a novel holistic approach for data exploration on dense knowledge graphs with a proof-of-concept in biomedical research. A knowledge graph (also known as semantic network) is highly related to the concept of an ontology which also holds representations of entities and their relations between them. We use both a more general and more specific definition of knowledge graph. It may contain multiple ontologies and inter-ontologie relations. But all entities need to be defined in a proper formal way with a formal semantic. We will discuss the data integration from multiple sources.% since not all relations are described in RDF or OBO format.

We propose the merging of knowledge graphs in general as a novel, holistic approach leading to  new applications in life sciences and graph algorithms. This process will build the foundation for quality control, validation of hypothesis, detection of missing data and time series analysis of biomedical wisdom in general. In addition this knowledge representation of linked data can be used to answer semantic questions.

Medical as well as biological and in general life-science researchers aim to understand -- for example -- the mechanisms of life, living organisms, and in general the underlying fundamental biological processes of life.  For example systems biology uses approaches like integrative knowledge graphs to decipher mechanism of a disease. In this field a lot of databases for disease modeling and pathways exist. In addition the Biological Expression Language (BEL, see \url{www.openbel.org}) is widely applied in biomedical domain to convert unstructured textual knowledge into a computable form. The BEL statements that form knowledge graphs are semantic triples that consist of concepts, functions and relationships \cite{fluck}. Thus they can be easily added to a knowledge graph. An example for a large Alzheimer network can be found in \cite{kodamullil2015computable}.

But biomedical information exists in more general form. Research is usually communicated using publications which are available in databases, for example MedLine and PubMed\footnote{See \url{https://www.ncbi.nlm.nih.gov/pubmed/}.}. These articles or abstracts are the source for biological relations mentioned above. In addition, meta information like authors, journals, keywords (so called MeSH-Terms, Medical Subject Headings), etc. are available. MeSH terms already provide a tree structure, forming a knowledge graph on their own. In addition, every ontology will form another knowledge graph. Ontologies are used as concepts in triples describing biological or medical relations. Using methods of natural language processing and text mining, we can combine and link these knowledge graphs to a giant and very dense new knowledge graph. This will meet a very general definition of \emph{context}. We can see every knowledge (sub-)graph as context to another. Biological expressions are context of the corresponding literature, authors are context of a text, named entities from ontologies found in a text are context to it or to the corresponding biological expressions.

At first we will describe the basic idea of this concept. The proof of concept will be done by applying this method to data from biomedical literature. We will also describe problems for data mining (missing and uncertain data) and especially the issues on scalable storage and for the analysis of this data.

%\begin{figure}[t]
%\centering
%\includegraphics[width=0.7\textwidth ]{dikw.pdf}
%\caption{ DIKW  hierarchy or `Knowledge Pyramid` in both a \emph{linear} and the \emph{pyramid} perspective.  }\label{abb:dikw}
%\end{figure}

\section{Preliminaries}

Data and Knowledge Management, sometimes also called Information Management, is a core topic of Data Engineering and Data Mining. %It is also a interdisciplinary field touching economics (how efficient and expensive is the solution?), psychology (do people use this solution in a way that was intended?) and of course informatics. This underlines the universality of this approach.
%
%One of the core concept is DIKW (Data, Information, Knowledge, Wisdom, see \cite{hey2004data}). First introduced by \cite{zeleny1987management} in 1987 it was developed by \cite{ackoff1989data} in 1989 who introduced the perspective of wisdom. Sometimes this hierarchy is depicted as a \emph{Knowledge Pyramid}, sometimes it is a linear chain. Figure \ref{abb:dikw} combines both perspectives: The linear perspective of understanding and context with past and future and the pyramid's perspective describing the amount of data leading to a smaller amount of information etc.
%In general, knowledge can be seen as explicit or implicit. Data is always given explicit. Implicit knowledge is not available for data mining, since it is only available as personal knowledge or experience.
%In information theory, knowledge is obtained from data and information. Data are recorded, context-free facts like measured values from devices (mass spectroscopy) or basic notes (weight of patients), but also images (e.g. computer tomography). The latter in turn may be regarded as organized pixels of certain intensity values. In the beginning, these data are measured. If this data is enriched by context, which implies meaning and purpose we get information. This information leads to knowledge and wisdom if -- once again -- enriched by context.

%More information about this topic can be found in the work of \cite{hey2004data} or \cite{doi:10.1177/0165551506070706}.
%
%We need to discuss the data to consider within our model.
The term \emph{terminology} is generally understood in relation to the SKOS meta-model \cite{SKOS} which can be summarized as Concepts unit of thoughts can be identified, labeled with lexical strings, assigned notations (lexical codes), documented with various types of note, linked to other concepts and  organized into informal hierarchies and association networks, aggregated, grouped into labeled and/or ordered collections, and mapped to concepts.

 Several complex models have been proposed in literature and have been implemented in software, see \cite{Zeng07}. \emph{Controlled Vocabularies} contain lists of entities which may be completed to a \emph{Synonym Ring} to control synonyms. \emph{Ontologies} also present properties and can establish associative relationships which can also be done by \emph{Thesauri} or \emph{Terminologies}. See \cite{NISOZ39} and \cite{Zeng08} for a complete list of all models.

 Here we define Terminologies similar to Thesauri as a set of concepts. They form a \emph{DAG} with child and parent concepts. In addition we have an associative relation which identifies similar or somehow related concepts. Each concept has one or more labels. One of them is the preferred identifier, all others are synonyms.

A \emph{Knowledge Graph} is a systematic way to connect information and data to knowledge. It is thus a crucial concept on the way to generate knowledge and wisdom, to search within data, information and knowledge. As described above, the context is the most important topic to generate knowledge or even wisdom. Thus, connecting Knowledge Graphs with context is a crucial feature.

We define knowledge graphs $G=(E,R)$ with entities $e\in E$ coming from a formal structure like a terminology $T$ or an ontology $O$. The relations $r\in R$ can be terminology or ontology relations, thus in general we can say every terminology $T$ or ontology $O$ which is part of the data model is a subgraph of $G$ which means $T\subseteq G$ or $O\subseteq G$. In addition we allow inter-terminology or inter-ontology relations between two nodes $e_1, e_2$ with $e_1 \in T_1$, $e_2 \in T_2$ and $T_1 \neq T_2$. More general we define $R=\{R_1,...,R_n\}$ as list of either terminologies or ontologies.

Both $E$ as well as $R$ are finite discrete spaces. Every entity $e\in E$ may have some additional meta information which need to be defined with respect to the application of the knowledge graph. The naming convention should follow Dublin Core, see \cite{weibel1997dublin}.

\section{Method}

% Das muss hier viel weniger theoretisch!

%Let $G=(E,R)$ be a Knowledge Graph with entities $E\subset \hat{E}$ in and relations $r\in R\subset\hat{R}$ between them. $\hat{E}$ as well as $\hat{R}$ are finite discrete spaces.

%Every entity $e\in \hat{E}$ may have some additional meta information which need to be defined with respect to the application of the knowledge graph. For instance there might be several ontologies $E_{1},...,E_{n}$ so that $E\subset \hat{E}= \cup_{i=1,...,n} E_{i}$. The same holds for $\hat{R}$.

\begin{figure*}[t]
	\begin{center}
		\includegraphics[width=0.65\textwidth]{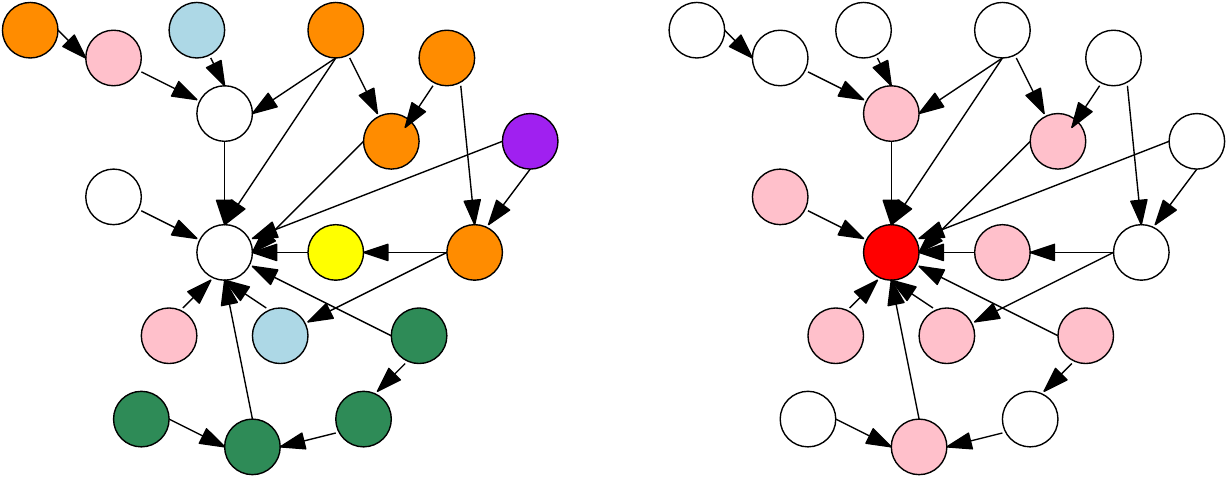}
		\caption{
			Illustration of a knowledge graph with several ontologies denoted by color (left). In the right illustration the context $con(e)$ of a red node $e$ is denoted by pink nodes.
		}
		\label{fig.context}
	\end{center}
\end{figure*}

We define contexts $C=\{c_{1},...,c_{m}\}$ as a finite, discrete set. Every node $v\in G$ and every edge $r\in R$ may have one ore more contexts $c\in C$ denoted by $con(v)$ or $con(r)$. It is also possible to set $con(v)=\emptyset$.  Thus we have a mapping $con:E\cup R\rightarrow \mathcal{P}(C)$. If we use a quite general approach towards context, we may set $C=E$. Thus every inter-ontology relation defines context of two entities, but also the relations within an ontology can be seen as context, see figure \ref{fig.context} for an illustration.

Thus here we assume that every information entity can also be a context information for other entities. For example a document can also be a context for other documents (e.g. by citing or referring to the other publication). An author is both a meta information to a document, but also itself context (by other publications, affiliations, co-author networks, ...). Other data is more obvious a context: named entities, topic maps, keywords, etc. extracted with text mining from documents. But already relations extracted from a text may stand for themselves, occurring in multiple documents and still valuable without the original textual information.

It is also possible to get the context of a subgraph $R_i\subseteq G$ which can be denominated by $con(R_i)$ or with the notation of graph theory as the extended induced subgraph by the vertex set $E_i$ from $R_i$ given by $G^c[E_i]$. This is quite trivial if context from $R_i$ can only be annotated to vertices in $G$. Then
\[G^c[E_i]=G[E_i] \cup \{ (e,e') \;\forall e'\in N(e), e\in E_i \} \]
If context can also be annotated to edges in $G$ this gets more complicated:
\[G^c[E_i]=G[E_i] \cup \{ (e,e') \;\forall e'\in N(e), e\in E_i \} \cup \{ (e',e'') \;\forall (e',e'')\in con_{|R}(e), e\in E_i \} \]
Here  $con_{|R}$ is the context of a node restricted to the set of edges (relations) in the graph. The two edges $e', e''$ are implicitly given by this context. It is quite easy to see that the restriction on context annotated to edges makes the problem more easy from a computational perspective. Nevertheless, context on edges is needed from a real-world perspective.

Having these knowledge graph $G$ with subgraphs $R_i$ it is also possible to add new relations. If two edges $e_1,e_2\in R_1$ are connected and $e'_1,e'_2\in R_2$ with $con(e_1)=e'_1$ and $con(e_2)=e'_2$ are not connected, we may add another edge $(e_1,e_2)$ with provenance information that this connection comes from a different context, namely $R_2$.

If the mapping $con$ is well defined for the domain set the Graph $G$ can be generated in polynomial time if all subgraphs $R_1,...,R_n$ are given. Since it is in general not the case that $con$ is already defined as a mapping, this usually contains a data or text mining task to generate contexts from free texts or knowledge graph entities. With respect to the notation described in \cite{dorpinghaus2018question} this problem $p$ can be formulated as
\begin{equation} \label{eq:1}
	p=\mathbb{D}|R|\mathbf{f}:\mathbb{D}\rightarrow\mathbb{X}|err|\emptyset
\end{equation}
Here, the domain set $\mathbb{D}$ is explicitly given by $\mathbb{D}=G$ or -- if additional full-texts $\hat{D}$ supporting the knowledge Graph $G$ exist -- $\mathbb{D}=\{G,\hat{D}\}$. In our case the domain subset $R=\mathbb{D}$. In this case we need to find a description function $f:\mathbb{D}\rightarrow\mathbb{X}$ with a description set $\mathbb{X}=C$ which holds all contexts. To find relevant contexts we need an error measure $err:\mathbb{D}\rightarrow[0,1]$.

\begin{figure*}[t]
	\begin{center}
		\includegraphics[width=0.35\textwidth]{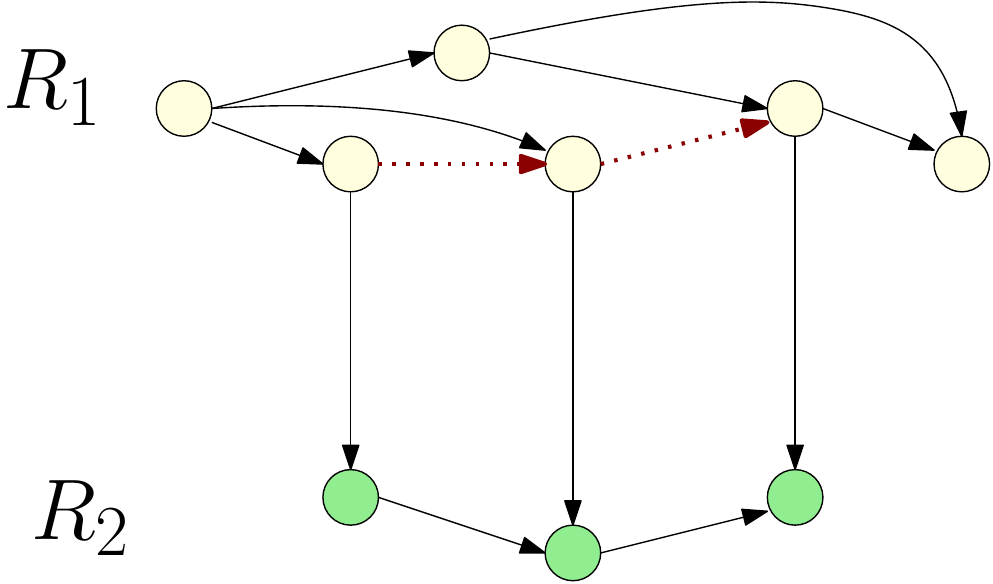}
		\caption{
			This figure describes the connection between two sets $R_1$ and $R_2$ in the knowledge graph $G$. With the information given by connections within $R_2$ it is possible to add new relations within $R_2$. The new edges are dotted.
		}
		\label{fig.metacon}
	\end{center}
\end{figure*}

It is also possible to build the context vice verse: Given a subgraph $G'=(E',R')\subseteq G=(E,R)$ the context graph is given by all contexts that are annotated to edges in $G'$. This can be either seen as inverse mapping $con^{-1}(G')$ or as the hypergraph $\mathcal{H}(G')=(X,\hat{E})$ given by
\[X=E'\cup G^c[E_i]\]
\[\hat{E}=\{ \{e_i, e \forall e\in N(e_i)\} \forall e_i \in X\}\]

\begin{figure*}[t]
	\begin{center}
		\includegraphics[width=0.45\textwidth]{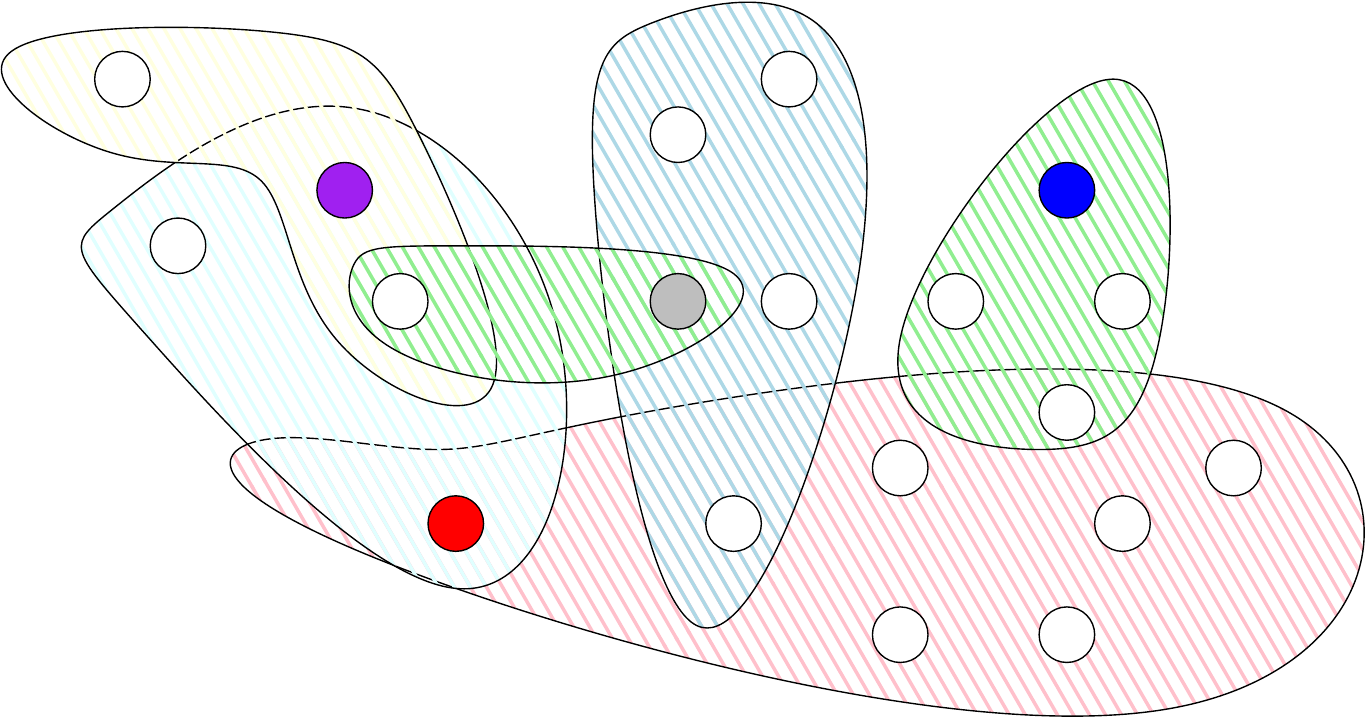}
		\caption{
			This figure illustrated the hypergraph $\mathcal{H}(G')=(X,\hat{E})$ for a subgraph $G'=(E',R')\subseteq G=(E,R)$. The colored nodes are additional context to the initial (white) nodes from $E'$. The hyperedges, illustrated by sets, connect nodes with context, but also nodes with the same context.
		}
		\label{fig.hypergraph}
	\end{center}
\end{figure*}

This graph can be seen as an extension of the original knowledge graph $G'$ where contexts connect not only to the initial nodes, but also every two nodes in $G'$ are connected by a hyperedge if they share the same context. See figure \ref{fig.hypergraph} for an illustration.

This graph $G=(E,R)$ and for two subset $R'\subseteq E$ or $G'\subseteq G$ the graphs $G^c[R']$ and $\mathcal{H}(G')=(X,\hat{E})$ can be used to answer several research questions and can be utilize to find their graph-theoretic formulations. In addition these graphs can be used as a basis for additional NLP and knowledge discovery. We will discuss a practical example for biomedical research.

\section{Dense knowledge graphs for biomedical research}

As a basis for building an initial knowledge graph for biomedical research we use two databases, MedLine and PubMed. PubMed contains 29 million abstracts from biomedical literature, PMC  about 4 million full-text articles.
The initial step of creating a document and context graph with basic context extraction needs a basic definition of entity sets $E_{1},...,E_{n}$ and their relations. Each relation stores a provenance and additional meta information.

The articles and abstracts from PubMed and PMC already store a lot of contextual data. We may set $E_{doc}$ as the document set containing nodes, each representation one document. In addition we may add a set $E_{origin}=\{\text{PubMed, PMC}\}$ as the origin of a document. Thus each document can be interpreted as context of a data source.

In addition we can store relevant meta information and set $E_{authors}$ as the set of authors, $E_{affiliation}$ as their affiliation which is again context for the authors. In addition we may set $E_{type}$ containing the article type. PubMed stores several classes, for example Books and Documents, Case Reports, Classical Article, Clinical Study, Clinical Trial, Journal Article, Review etc. These entity classes are neither ontologies nor terminologies but can be interpreted as such. We can see, that the context already leads to novel relations. $E_{affiliation}$ will lead to edges like \texttt{sameAffiliation} in  $E_{authors}$, $E_{doc}$ to \texttt{isCoAuthor}.

Another important context is $E_{mesh}$ storing the keywords, which come from the MeSH (Medical Subject Headings) tree, see \cite{rogers1963medical} and \url{https://www.nlm.nih.gov/mesh/intro_trees.html}. Thus $E_{mesh}$ already comes with a hierarchy and edges $R_{mesh}$. This pushes us nearer to the initial idea of an ontology.
The value of MeSH terms and their hierarchy for knowledge extraction was shown in several recent studies like \cite{yang2018research}.

\begin{figure*}[t]
	\begin{center}
		\includegraphics[width=0.6\textwidth]{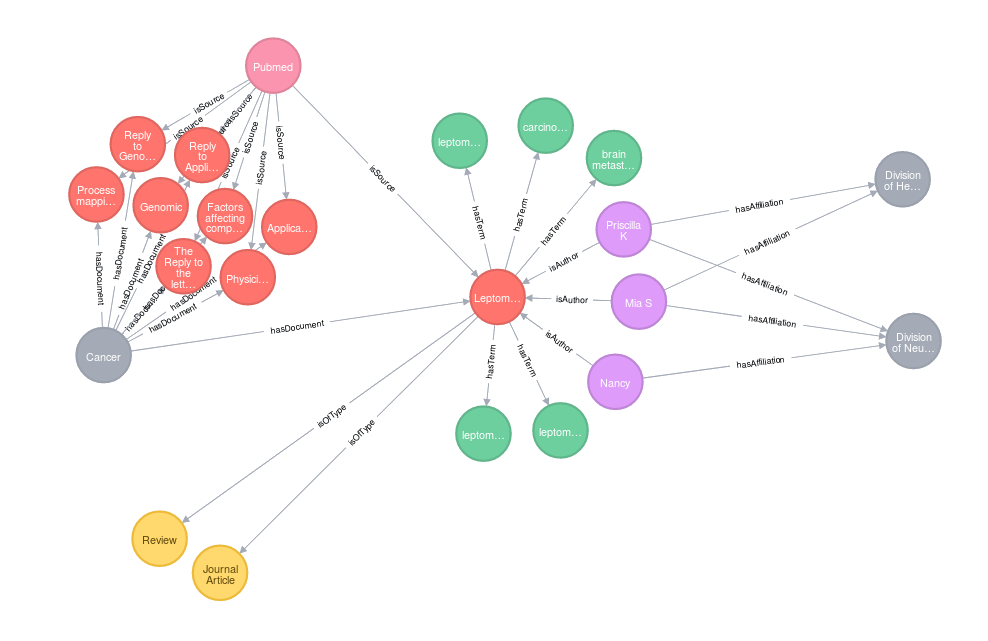}
		\caption{
			This figure is an illustration of the initial document and context graph. A "Pubmed" node is the source of document nodes (red). There are several context annotations like article type (yellow), keywords (green), authors (purple). Authors have additional context (affiliations, gray).
		}
		\label{fig.context1}
	\end{center}
\end{figure*}

All other relations can be added between the sets $E_{i}$, for example $R_{coauthors}$, $R_{hasAffiliation}$, etc. With these information given it is -- from an algorithmic point of view -- quite easy to add all context relations like $R_{isAuthor}$, $R_{hasDocument}$, $R_{hasAnnotation}$ etc. See figure \ref{fig.context1} for an illustration.

The initial knowledge graph can be extended by NLP-technologies using ontologies or terminologies for named entity recognition (NER). This will result in a hierarchy within this ontology. For example the Alzheimer's Disease Ontology (ADO, see \cite{MALHOTRA2014238}) $E_{ADO}$ or the Neuro-Image Terminology (NIFT, see \cite{iyappan2017neuroimaging}) $E_{NIFT}$ coming with their hierarchy $R_{ADO}$, $R_{NIFT}$ can be used. The process of NER will lead to another context relation $E_{hasAnnotation}$. Since not all ontologies or teminologies are described in RDF or OBO format we have to add data from multiple sources. This is done by a central tool providing all ontology data.

Relation extraction on literature can also be used to extend the knowledge graph since it is based on NER. For example we can use the state-of-the-art, BEL-based, technologie from BELIEF-workflow, see \cite{belief2}. This will add an additional edge-set $R_{BEL}$ based on the BEL specification.

Another context data useful for knowledge extraction are citations, thus edges $R_{hasCitation}$ between two nodes in $E_{doc}$. The data from PMC already stores citation data with unique identifiers (PubMed IDs). Some data is available with WikiData, see \cite{voss2016classification} and \cite{VrandecicDL2018}. Other sources are rare, but exist, see \cite{osswald2015continuing}. Especially for PubMed a lot of research is working on this difficult topic, see for example \cite{volanakis2018sciride}.

\section{Technical Design, Implementation and Performance Evaluation}

The technical design was done with respect to the microservice architecture of SCAIView, see \cite{scaiview}. The service \texttt{scaiview-graphstore} has both a REST API as well as a JMS interface to communicate with the microservice ecosystem and to retrieve data which will be stored as a graph. Here, Spring Data maps objects to graphs. As a database backend, we used Neo4J. Thus our software can be used to perform Cypher and SPARQL queries. Data can be retrieved in JSON or RDF format. Thus, this service provides FAIR (meta-)data by design. The data is findable, accessible, interoperable (due to the usage of controlled vocabulary) and reusable.

Unfortunately, we were unable to store the complete knowledge graph within the database. The performance was rather disappointing. This was probably as a result of the fact that the knowledge graph is very dense and thus not very scalable.

%\section{Performance Evaluation}

\section{Data Validation and Exploration using MDDs}

Validation of hypothesis or fact checking on networks is a widely considered issue, see \cite{ciampaglia2015computational} or \cite{shiralkar2017finding}. Most popular is the shortest path method between two entities. Shiralkar et al. introduced the concept of \emph{finding} or \emph{knowledge streams} which consider all relevant paths between two entities. Here, we propose a more general approach using multiple-valued decision diagrams (MDDs).

Multiple-valued decision diagrams (MDDs) are an efficient data structure that in several fields like verification, optimization and dynamic programming have become the method of choice for representing a logic function or a model of constraints.
MDDs are also implemented in many dynamic programming solvers. See, e.g.\ \cite{andersen}, \cite{hadzic}, \cite{perez2015efficient}, \cite{bergman2011manipulating}.
In the last decade, MDDs have been studied extensively. For a survey, see \cite{bergman}. One of the advantages of MDDs is their high potential of compression and that the basic operations to modify an MDD can be implemented very efficiently. Recently, efficient operators such as union, negation, intersection, etc.\ (see, e.g.\ \cite{perez2015efficient}) as well as parallelized algorithms for compression (\cite{perez2018}) have been designed.

In this paper, we only consider reduced and ordered MDDs. They are a generalization of reduced and ordered binary decision diagrams which where introduced in \cite{bryant}.
We call an (multiple-valued or binary) decision diagram reduced if redundant nodes are eliminated and isomorphic nodes are merged (cf.\ \cite{bryant}).
Furthermore, ordered MDDs use a fixed variable ordering for canonical representation. An MDD is a rooted directed acyclic graph (DAG) used to represent some binary function $f: (x_1,\dots, x_k)\mapsto \{\text{true}, \text{false}\}$ where $x_i\in\{1,\dots,d_i\}$ for $i=1,\dots,k$. Given the $k$ input variables, the MDD contains $k+1$ layers of nodes, such that each variable is represented at a specific layer of the graph.
The additional layer consists of the two terminal nodes \textit{true} and \textit{false}. Every node on layer $i=1,\dots,k$ has at most $d_i$ outgoing edges to nodes in the next layer of the graph. Every outgoing edge represents one of the possible values in the domain of the variable.

\begin{figure*}[t]
	\begin{center}
		\includegraphics[width=0.85\textwidth]{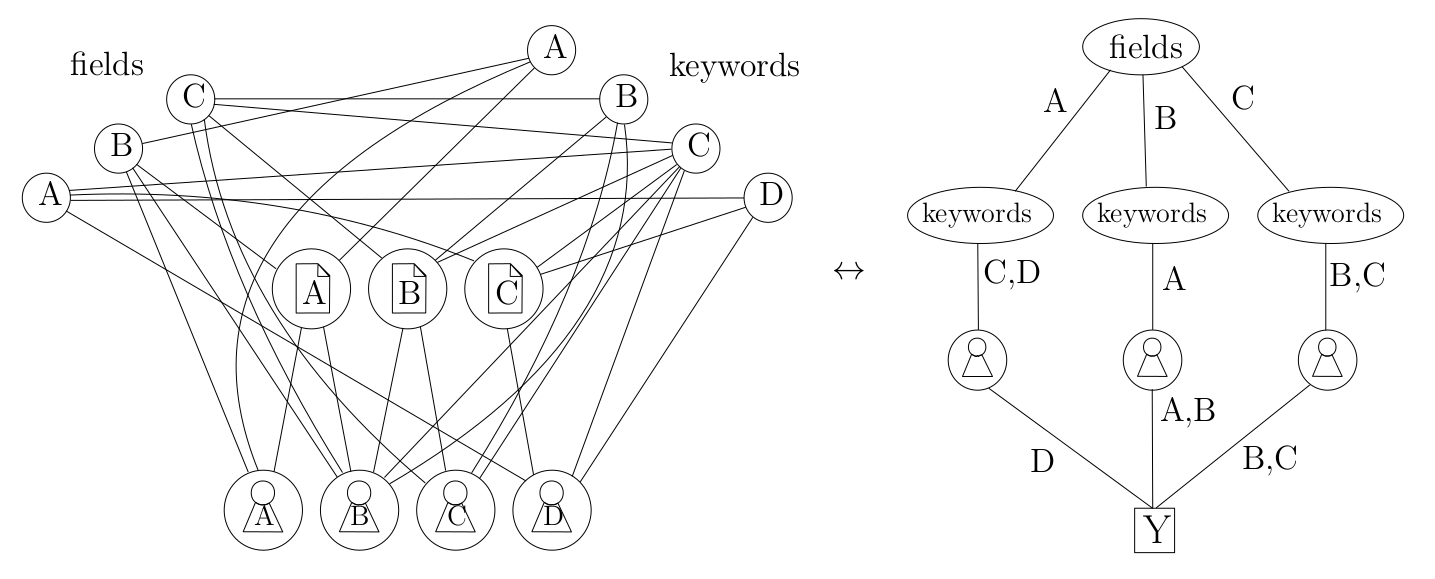}
		\caption{
			A knowledge graph (left) with research fields, keywords, documents and authors. A MDD (right) starting with the fields will describe all possible combinations and flows
		}
		\label{fig.mdd1}
	\end{center}
\end{figure*}

A simple example for a transformation from a knowledge graph to a MDD can be found in figure \ref{fig.mdd1}.
The knowledge graph represents documents and their context: research fields, keywords and authors. The MDD is rooted in the research fields.
This MDD spans the field of all possible combinations within the graph. The terminal node \textit{true} is labeled with ``Y''.
To keep things clear, the terminal node \textit{false} as well as all paths that correspond to non-existing combinations of research fields, keywords and authors, and therefore end in the \textit{false} node, are omitted.
In general, this MDD should either be limited to a certain depth or both root-node and end-node need to be set.

\begin{figure*}[t]
	\begin{center}
		\includegraphics[width=0.4\textwidth]{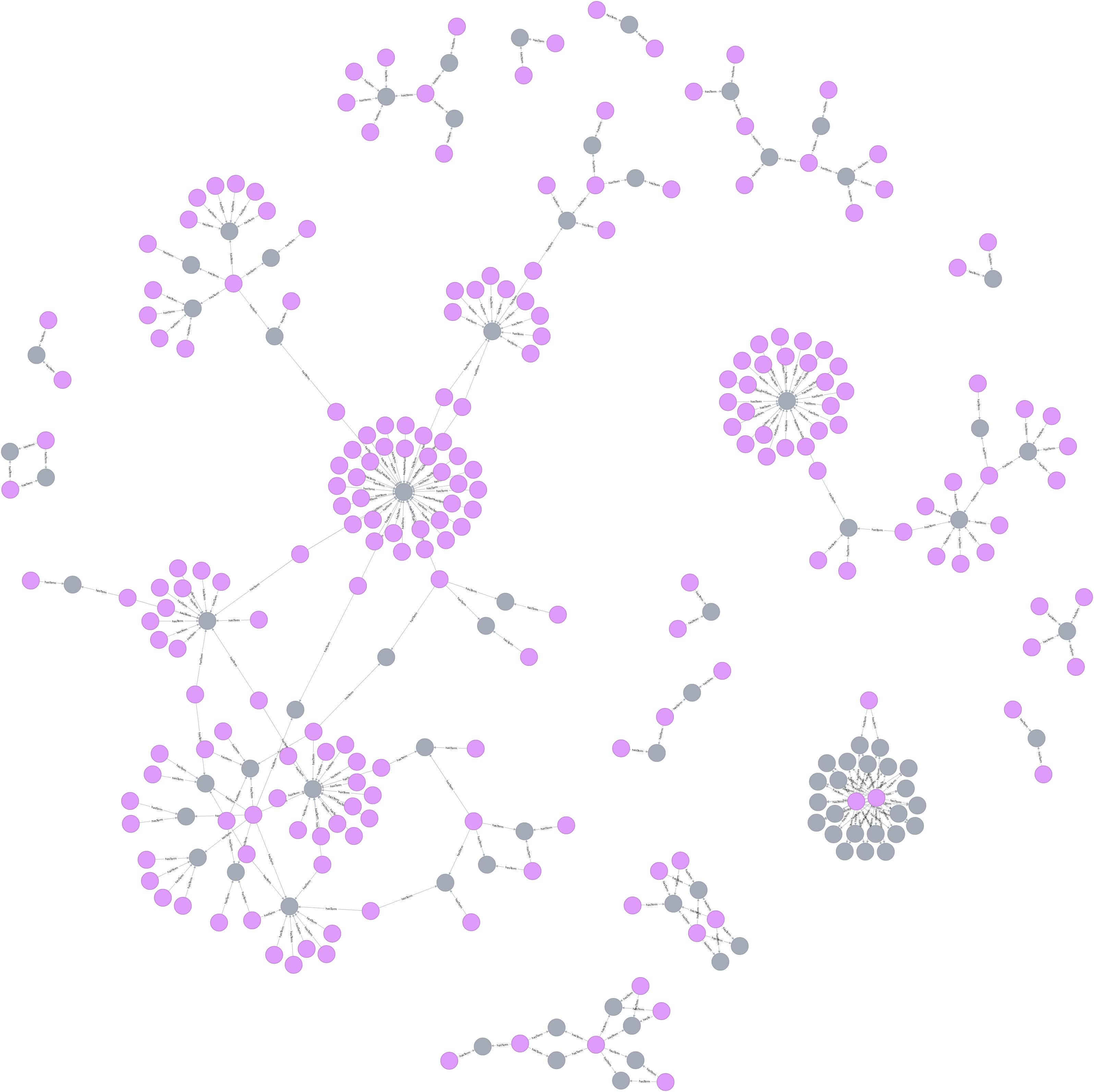}
		\quad
		\includegraphics[width=0.4\textwidth]{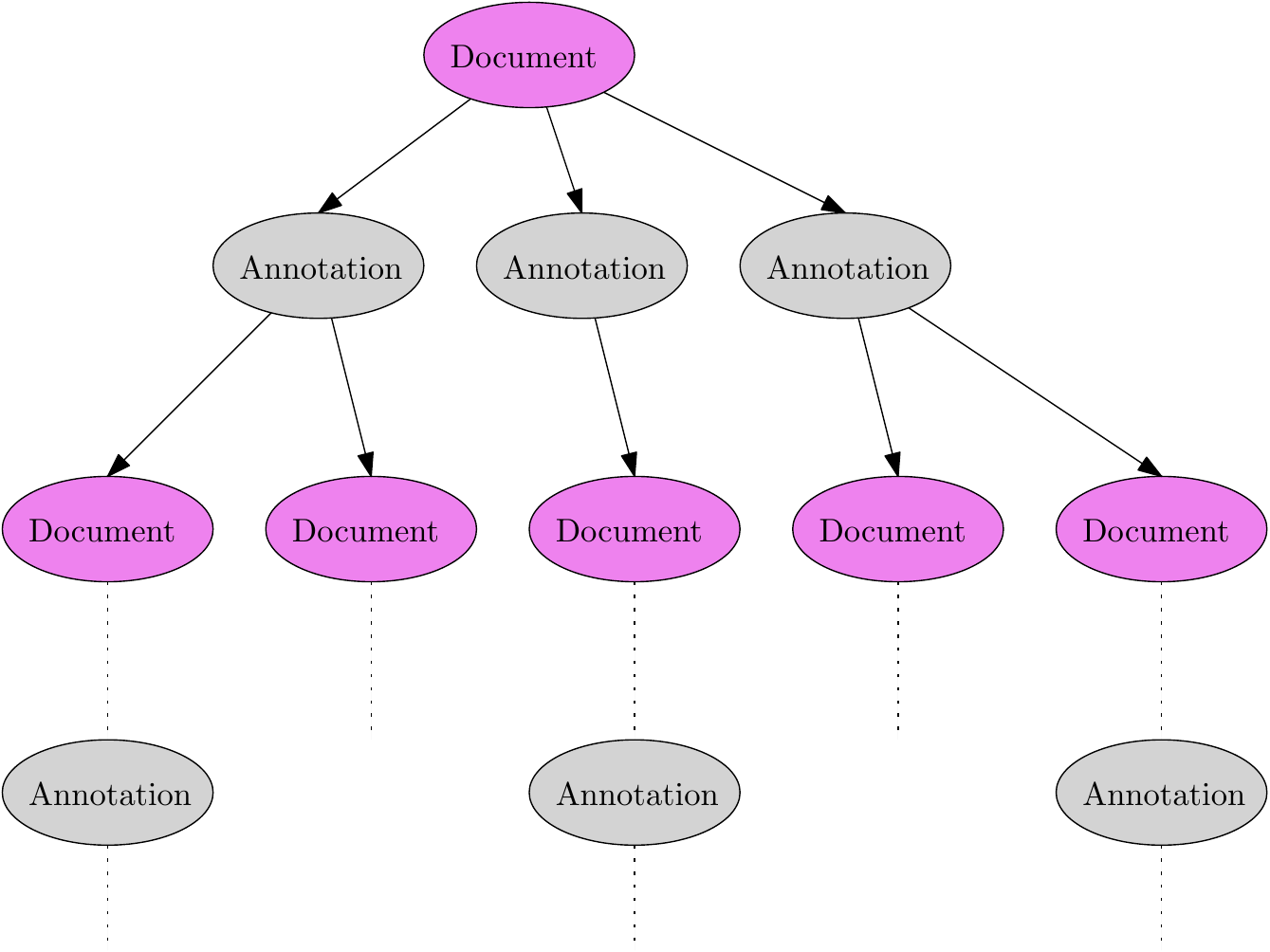}
		\caption{
			On the left the knowledge graph restricted to documents (purple) and annotations (gray). On the right the corresponding MDD with alternating document and annotation layers. Each decision on document and annotation will lead to a different set of annotations and documents which can be used for knowledge discovery.}
		\label{fig.mdd3}
	\end{center}
\end{figure*}

In the first case, each decision step will uncover a new layer of either keywords or documents. This can be seen as an implementation of manual creation of topic maps. Keywords are used to search for documents that enrich a corresponding topic in the knowledge graph. Same or similar sets of kewords usually describe the same topics or topics close to that. The non-overlapping fraction of keywords guides to new topics and new documents going along with that. In summary, you have a knowledge assembly consisting of literature in form of clusters of documents each illuminating a topic. In the knowledge graph, these documents are linked via annotations, a pool of different entities from different sources, that describe the content features of the document on the whole. Keywords are one feature category. In the map, interlinked document clusters can either display a step to an increased granularity in the knowledge stream and describe a topic in more detail or they open up a new route to new contexts thereby connecting research foci. Such a knowledge map can be converted into a MDD because it fulfils corresponding criteria: 1) a finite amount of documents in the knowledge map 2)  documents or annotations can be defined as  ``source'' 3) decisions can be made between annotations and guide to new documents and vice versa 4) each route has a  ``sink'' 5) annotations or documents shared between different decision routes can be merged to remove redundant nodes/route-sections. The resulting MDD collection of ``decision routes'' may help to refine the knowledge map by adding new contexts or even new topics. See figure \ref{fig.mdd3} for an illustration.%See figure \ref{fig.mdd3} for an illustration.

\begin{figure*}[t]
	\begin{center}
		\includegraphics[width=0.4\textwidth]{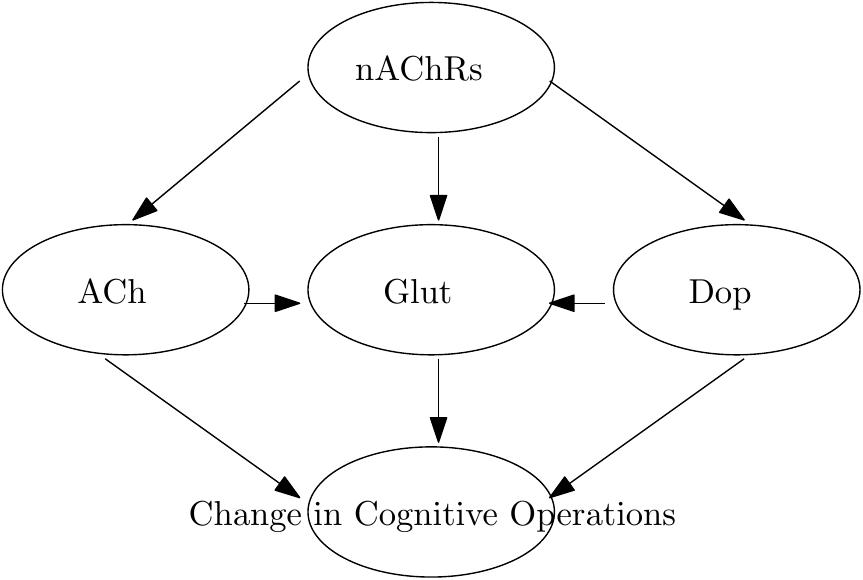}
		\quad
		\includegraphics[width=0.4\textwidth]{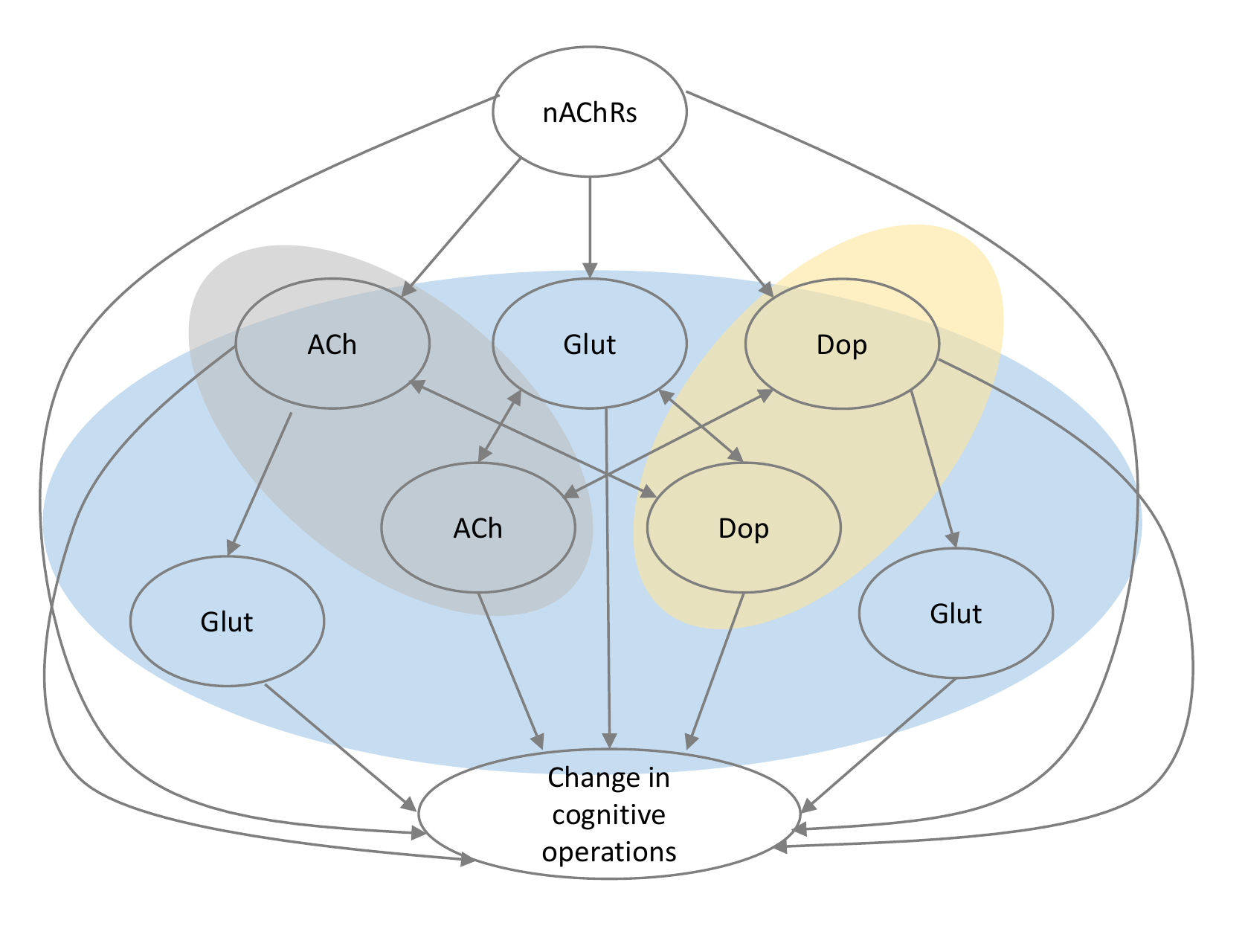}
		\caption{
			A knowledge graph (left) describing the interaction of nicotinic acetylcholine receptors with the cholinergic, glutamatergic and dopaminergic system. The action of nicotinic receptors can influence the activity in neurotransmitter systems depending on whether they are active or not. Different combinations of nACh receptor activity and transmitter system activities form distinct routes that result in changes of cogitive operations.
	 		A MDD (right) starting with the receptor layer will result in changed cognitive operations or no change. %The green edge is forbidden since a MDD should be acyclic, but for a proper representation of the knowledge graph they need to be added. The red and green edges extend the MDD to a finite state machine.
		}
		\label{fig.mdd2}
	\end{center}
\end{figure*}

In the second case, having both a fixed  root-node $a$ and end-node $b$, this can be used to validate the knowledge of the influence of $a$ on $b$. This MDD will contain all paths from node $a$ to $b$ in $G$ and their logical sequence. This is both: a valid help for the manual curating person and a basis for further algorithmic consideration.

We need to discuss another example that will show the relations between multiple-valued decision diagrams, finite state machine and Markov chains, see the works of \cite{hermanns1999multi} and \cite{shrestha2010decision}. The biological knowledge graph shown here deals with the nicotinic acetylcholine receptor (nAChR) mediated control of learning and memory pathways. These receptors are ligand-gated pentameric ion channels, belong to the cys-loop receptor family, are ubiquitously expressed in the brain and equipped with various regulatory functions \cite{dani2007nicotinic}. Nicotinic subunits can either form homomeric receptors (only alpha subunit) or hereromeric receptors consisting of a mixture of alpha and beta subunits with varying stoichiometry \cite{millar2009diversity}. They get activated by the endogenous ligand acetylcholine or via corresponding agonists such as nicotine \cite{dani2015neuronal}. Nicotinic receptors can regulate several neurotransmitter systems \cite{placzek2009nicotinic} \cite{cao2005nicotinic} and be part of biological pathways that lead to the induction of synaptic plasticity which in turn can results in altered cognitive operations \cite{yakel2014nicotinic}\cite{newhouse2001nicotinic}. Our knowledge graph depicts a cut-out of a giant nAChR interaction network (data not shown). Here we focus on the cholinergic, glutamatergic and dopaminergic system. In our model, nAChR activity is supposed to cause change in cognitive operations. Functionally, this can occur via presynaptic modulation of transmitter release, or via postsynaptic modification of a neurons response to a given stimulus. Regulatory action of nAChRs in combination with one or more active transmitter systems describes possible routes of neuronal information processing with a change in cognitive operations as output. The MDD representation of this knowledge graph combines redundant nodes (transmitter systems that appear several times) for simplification purposes. Decision routes in the graph are formed based on whether the question ``Is the biological entity (receptor/ transmitter system) active?'' answered with  ``yes'' or  ``no'' for the corresponding node. These steps result in a finite number of potential routes of information transmission that have an impact on cognition. These MDD pathways can be compared with e.g. BEL-endcoded pathway information retrieved from example related literature. Matches serve to support existing knowledge and additional MDD-pathway could serve to make predictions about alternative biological pathways which in turn could be tested with biological assays.
See figure \ref{fig.mdd2} for an illustration.

For the purpose of validating a network it is also possible to add negative results, thus plasticity can be true or false. It is quite easy to see that this model can be interpreted and extended to a finite state machine. This can be done by adding the conflicting edges that were removed due to the acyclic property of MDDs.

\section{Conclusion}

The results of this study support the idea of a more holistic approach towards dense knowledge graphs. Our research and the discussed a proof-of-concept on a biomedical knowledge graph combining several sources of data as context suggest that it is possible to build and query these graphs. We processed data from PubMed and PMC. This initial knowledge graph was extended with state-of-the-art text mining and NER. Thus we were able to provide both small datasets as well as large collections of data. We were able to use semantic queries using Cypher or SPARQL to retrieve data. The data is FAIR by construction.

The most important limitation lies in the database infrastructure. Thus, this picture is still incomplete. We need to consider more general data-warehousing solutions and test other graph databases like cray graph engine.

We have devised a strategy which uses multiple-valued decision diagrams for data validation and exploration. We suggested both a ``decision-route'' to explore new data as well as a transformation of knowledge into MDDs which can be extended to a finite state machine. Our study provides the backbone for further research on algorithms and simulation.

%We will first of all discuss some missing data or data integration problems as well as technical issues which need to be solved. Afterwards we will give an outlook on NLP based on context information and the impact on answering semantic questions. This is highly related to the FAIRification of research data. This will lead to a short outlook on personalised medicine.

%%% Angabe der .bib-Datei (ohne Endung) / State .bib file (for BibTeX usage)
%\bibliography{lit}
%\printbibliography %if you use biblatex/Biber
%\printbibliography[heading=bibintoc,title={References}]

\bibliographystyle{IEEEtran}
\bibliography{lit}

\end{document}